\documentclass[runningheads]{llncs}
\usepackage{graphicx}
\usepackage{siunitx}
\usepackage{amsmath}
\usepackage{tabularx}
\usepackage{xcolor}
\usepackage{longtable}
\usepackage{booktabs,colortbl, array}
\usepackage{pgfplotstable}
\usepackage{tabu}
\usepackage{multirow}
\usepackage[utf8]{inputenc}
\usepackage{dblfloatfix}
\usepackage{float}
\usepackage{hyperref}
\usepackage[misc,geometry]{ifsym} 

\begin{document}
\title{Geometric Deep Learning for the Assessment of Thrombosis Risk in the Left Atrial Appendage}

%
\titlerunning{Geometric DL for the assesment of Thrombosis Risk in the LAA}
%

\author{ }

\author{Xabier Morales\inst{1}\textsuperscript{(\Letter)}  \and
Jordi Mill\inst{1}\and
Guillem Simeon\inst{1}\and
Kristine A. Juhl\inst{2} \and\newline
Ole De Backer \inst{3} \and
Rasmus R. Paulsen\inst{2} \and
Oscar Camara\inst{1}
}

\authorrunning{X. Morales et al.}

\institute{Physense, BCN Medtech, Department of Information and Communications Technologies, Universitat Pompeu Fabra, Barcelona, Spain \newline
\email{xabier.morales@upf.edu}\and 
DTU Compute, Technical University of Denmark, Kongens Lyngby, Denmark. \and 
Heart Center, Rigshospitalet, Copenhagen, Denmark.
}

\maketitle              
\begin{abstract}

The assessment of left atrial appendage (LAA) thrombogenesis has experienced major advances with the adoption of patient-specific computational fluid dynamics (CFD) simulations. Nonetheless, due to the vast computational resources and long execution times required by fluid dynamics solvers, there is an ever-growing body of work aiming to develop surrogate models of fluid flow simulations based on neural networks. The present study builds on this foundation by developing a deep learning (DL) framework capable of predicting the endothelial cell activation potential (ECAP), linked to the risk of thrombosis, solely from the patient-specific LAA geometry. To this end, we leveraged recent advancements in Geometric DL, which seamlessly extend the unparalleled potential of convolutional neural networks (CNN), to non-Euclidean data such as meshes. The model was trained with a dataset combining 202 synthetic and 54 real LAA, predicting the ECAP distributions instantaneously, with an average mean absolute error of 0.563. Moreover, the resulting framework manages to predict the anatomical features related to higher ECAP values even when trained exclusively on synthetic cases.

\end{abstract}
%
%
%

\section{Introduction}
\setcounter{footnote}{0}

Atrial fibrillation (AF) is the most common clinically significant arrhythmia, which can lead to irregular contraction and wall rigidity of the left atrium (LA). This often results in atrial blood stagnation promoting the formation of thrombi within the LA, thereby, increasing the risk of cerebrovascular accidents \cite{Watson2009}. In fact, non-valvular AF is responsible for 15 to 20\% of all cardioembolic ischemic strokes, 99\% of which originate in the left atrial appendage (LAA) \cite{Cresti2019}. As a result, there have been several attempts at characterizing LA haemodynamics either through transesophageal echocardiography (TEE) or computational fluid dynamics (CFD). Yet, ultrasound imaging is quite ill-suited to characterize complex three-dimensional haemodynamics, while the latter suffers from tediously long computing times and demands huge computational resources \cite{Liang2018}. 

In this regard, deep learning (DL) has made its way into fluid flow modelling, resulting in highly accurate surrogate models that can be evaluated with significantly less computational resources \cite{Hennigh2017}. That being said, many of the most widespread DL models are not well adapted to non-Euclidean domains, such as graphs and meshes, in which medical data is often best represented \cite{Fey_2018_CVPR}. As a response, a set of methods have emerged under the umbrella term Geometric DL, that have succeeded in generalizing models such as convolutional neural networks (CNN) to non-Euclidean data \cite{GeoDL}.

Hence, seeking to improve upon prior studies \cite{Morales2020}, we leveraged Geometric DL to develop a CFD surrogate capable of learning the complex relationship between the heterogeneous LAA geometry and the endothelial cell activation potential, parameter linked to an increased thrombosis risk. More specifically, we employed a spline-based spatial convolution operator, which enables extracting features from the underlying anatomy without the need for mesh correspondence \cite{Fey_2018_CVPR}, i.e., they extend properties that have made classical CNNs so successful (local connectivity, weight sharing and shift invariance), without the need to convert LAA meshes to Euclidean representation. We show that our model not only is accurate, but also generalizes well from synthetic to real patient data.


\section{Methods}

\begin{figure}[!]
\includegraphics[width=1\textwidth]{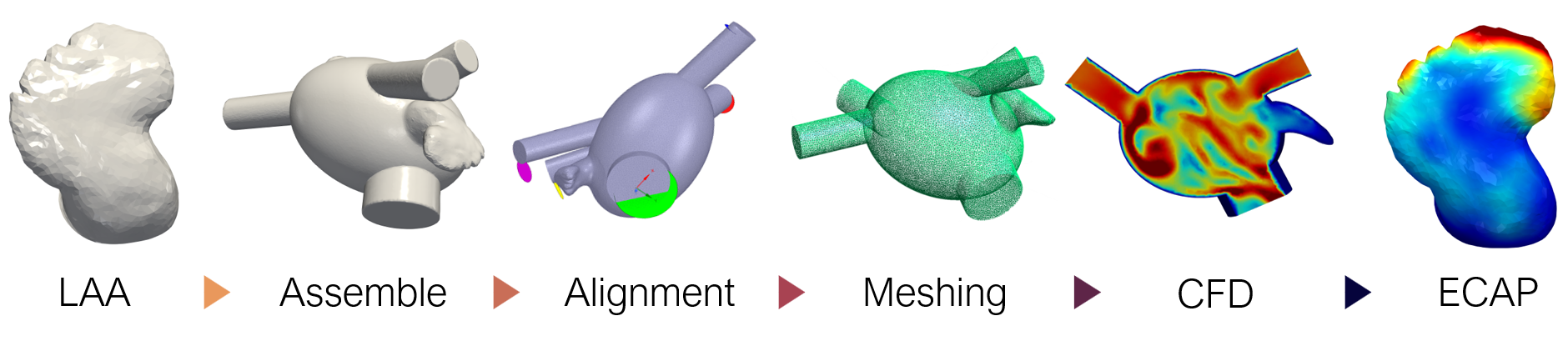}
\caption{Pipeline to generate ground truth ECAP maps. LAA: Left Atrial Appendage, CFD: Computational Fluid Dynamics, ECAP: Endothelial Cell Activation Potential.}
\label{fig:General Pipeline}
\end{figure}

The pipeline of the study involved, at first, the generation of the ground truth data through in-silico CFD simulations of the entire LA, requiring prior assembly of the geometries as shown in Figure \ref{fig:General Pipeline}. Subsequently, the meshes derived from the simulations were converted to graph format, suitable for the training of the geometric neural network. Finally, the model was trained seeking to learn the complex non-linear relationship between the LAA anatomy and the ECAP maps.

\subsection{Data}

The employed dataset consisted of 256 LAA, combining 202 synthetic and 54 real patient geometries. The synthetic geometries and their corresponding simulations were borrowed from a previous study \cite{Morales2020}. More specifically, the synthetic dataset stems from a statistical shape model (SSM) based on 103 patient LAA surfaces \cite{Slipsager2019}. All cases were reconstructed from computed tomography (CT) images provided by the Department of Radiology of Rigshospitalet, Copenhagen.

For the time being, we have just considered the geometry of the LAA as incorporating the highly heterogeneous LA anatomy would qualitatively increase the inter-subject variability of the hemodynamic parameters. Thus, prior to the simulations, all LAA were assembled to an oval approximation of the LA \cite{Garcia-Isla2018} to ensure that ECAP variability solely depended on individual anatomical differences of the appendage. Finally, since the employed framework does not require any sort of mesh correspondence, all the synthetic data were remeshed to ensure that the network was only able to learn from geometric features.

\subsection{ In-silico thrombosis risk index - ECAP}

The endothelial cell activation potential (ECAP), defined by Di Achille et al. \cite{DiAchille2014}, was the parameter chosen to evaluate the risk of thrombosis in the LAA. Since the pathophysiology of thromboembolism in AF is based upon the formation of mural thrombi, the calculation of ECAP is based upon haemodynamics in the proximity of the vessel wall, more precisely, as the ratio between the oscillatory shear index (OSI) and the time averaged wall shear stress (TAWSS). 

\begin{equation}
    ECAP= \dfrac{OSI}{TAWSS}
\end{equation}

High ECAP values result from low TAWSS and high OSI values, indicating the presence of low velocities and high flow complexity, which is associated with endothelial susceptibility and risk of thrombus formation. The ground truth ECAP distributions were obtained through CFD simulations performed on Ansys Fluent 19.2\footnote{https://www.ansys.com/products/fluids/ansys-fluent} and automated through the MATLAB R2018b Academic license\footnote{https://es.mathworks.com/products/matlab.html}. Simulation setup was performed accordingly to the preceding study \cite{Morales2020}.

\subsection{Geometric deep learning framework}

The model was constructed by leveraging PyTorch Geometric (PyG)\footnote{https://github.com/rusty1s/pytorch\_geometric}, a Geometric DL extension of PyTorch\footnote{https://pytorch.org/}. PyG offers a broad set of convolution and pooling operations that extend the capabilities of traditional CNN to irregularly structured data such as graphs and manifolds. With this in mind, the mesh dataset resulting from the simulations had to be converted into individual graphs. Together with PyVista\footnote{https://docs.pyvista.org/}, we converted each mesh to a graph represented by $\mathcal{G} = (\mathcal{V}, \mathcal{E})$, with $\mathcal{V} = {1, . . . , N}$ being the set of nodes, and $\mathcal{E}$ corresponds to the set of edges of the triangular faces. For each vertex the curvature and surface normal vectors were computed, totaling 4 input feature channels.

Among all the available graph CNN layers, we opted for SplineCNN  \cite{Fey_2018_CVPR}, since being a spatial method, it offers several advantages when dealing with meshes. In particular, it avoids the need of establishing mesh correspondence. Additionally, defining the spatial relations between vertex features becomes trivial by employing pseudo-coordinates. In our use case, pseudo-coordinates were obtained by computing the relative distance in Cartesian coordinates between the vertices of each edge. During the training process, these edge attributes define the way in which the input features will be aggregated in the neighborhood of a given node. Lastly, we also tested the residual and dense layers for graph neural networks developed by Li et al. \cite{li2020deepergcn}, aiming to reduce vanishing gradients in deep layers.

\subsection{Experimental setup and hyperparameter tuning}

\begin{figure}[!t]
\includegraphics[width=1\textwidth]{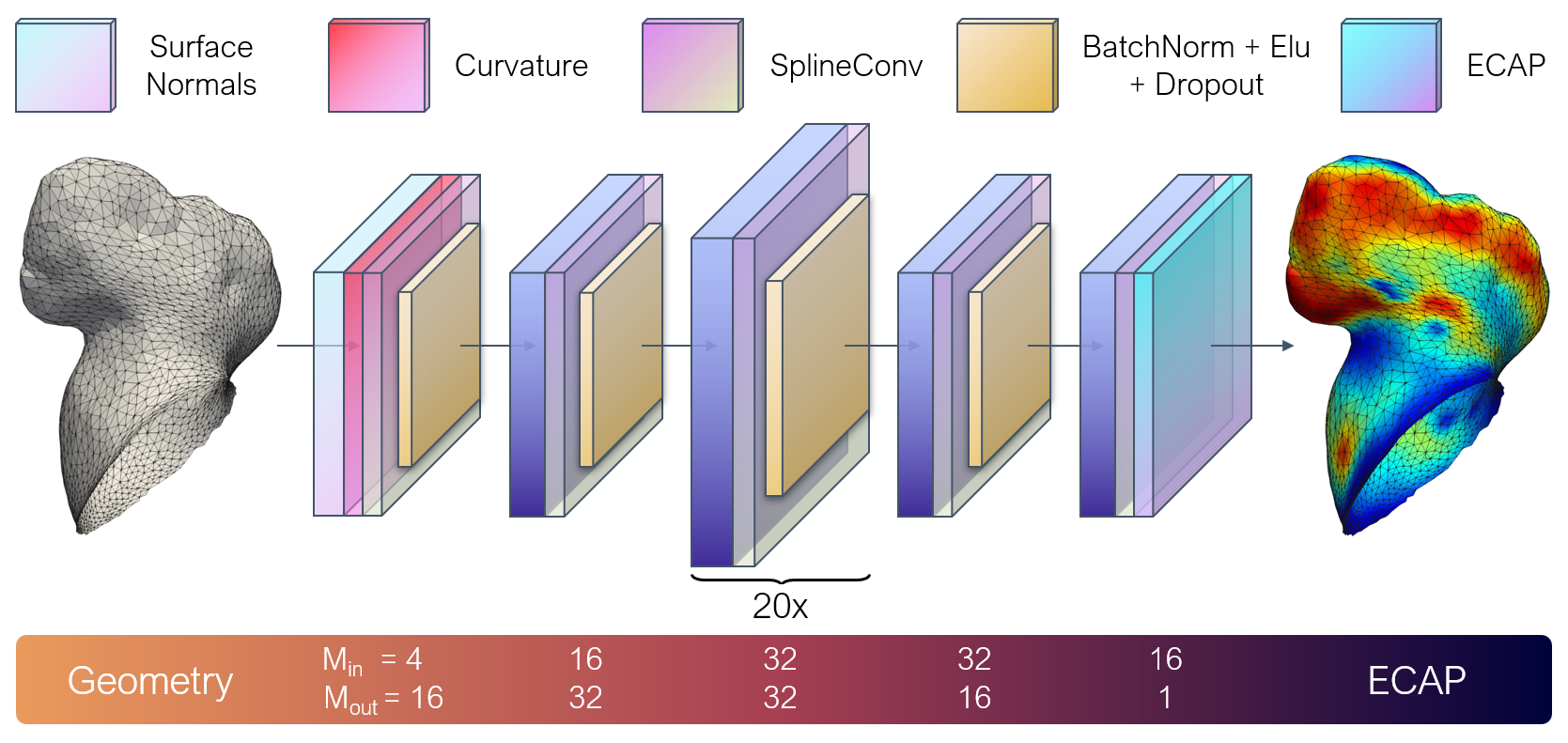}
\caption{General overview of the network architecture. The input vertex features consisted on the vertex-wise curvature and normal vectors. The spatial relations between the nodes were stored as edge attributes through Cartesian pseudo-coordinates. We employed 24 consecutive SplineConv layers  \cite{Fey_2018_CVPR} with a kernel size of $k=5$.}
\label{fig:Model architecture}

\end{figure}

The schematic representation of the model architecture is shown in Figure \ref{fig:Model architecture}. A thorough grid search was carried out to fine tune the model by iteratively swapping several hyperparameters, sequentially increasing model depth from 5 to 25 layers and including dense blocks of different sizes with a fixed random seed. The highest accuracy was obtained when employing 20 consecutive SplineConv hidden layers with 32 feature channels per layer. Besides, the ideal amount of transition layers from input-output to the hidden layers was also tested. The inclusion of one transition layer in both ends, as observed in Figure \ref{fig:Model architecture}, yielded the best performance. Moreover, several configurations of residual connections were evaluated, with the best results attained using dense blocks of depth $= 4$. Various pooling and U-Net like models were tested, aiming to improve multi-scale feature extraction, but so far to no avail.

In regards to the parameters of the SplineConv layer, a B-spline basis of degree 1 and a kernel size of $k=5$ were chosen, following suggestions by the authors \cite{Fey_2018_CVPR}. Concerning general hyperparameters, the exponential linear unit (ELU) provided the best results among all activation functions, always coupled with batch normalization and a dropout of 0.1. In addition, the training loop was carried out through $300$ epochs with a batch size of $16$ and a learning rate of $0.001$. Adam was employed as an optimizer with a weight decay of 0.05 when training in synthetic only. Finally, the L\textsubscript{1} loss was chosen for regression.

Given the limited availability of comparable models in similar tasks, the performance of the model was benchmarked against one of our earlier studies \cite{Morales2020}. As opposed to the novel graph-based network, this study relied on conventional fully connected layers (FCN) and therefore it required thorough preprocessing of the input meshes. Two separate experiments were completed. In the first, a 10-fold cross-validation was performed with the whole dataset, meaning that the model was given both synthetic and patient data during training and testing. In the latter, we trained the model solely on synthetic data and tested the accuracy of the model on the real cases to test its generalization capabilities.


\section{Results}

\begin{figure}[H]
\includegraphics[width=1\textwidth]{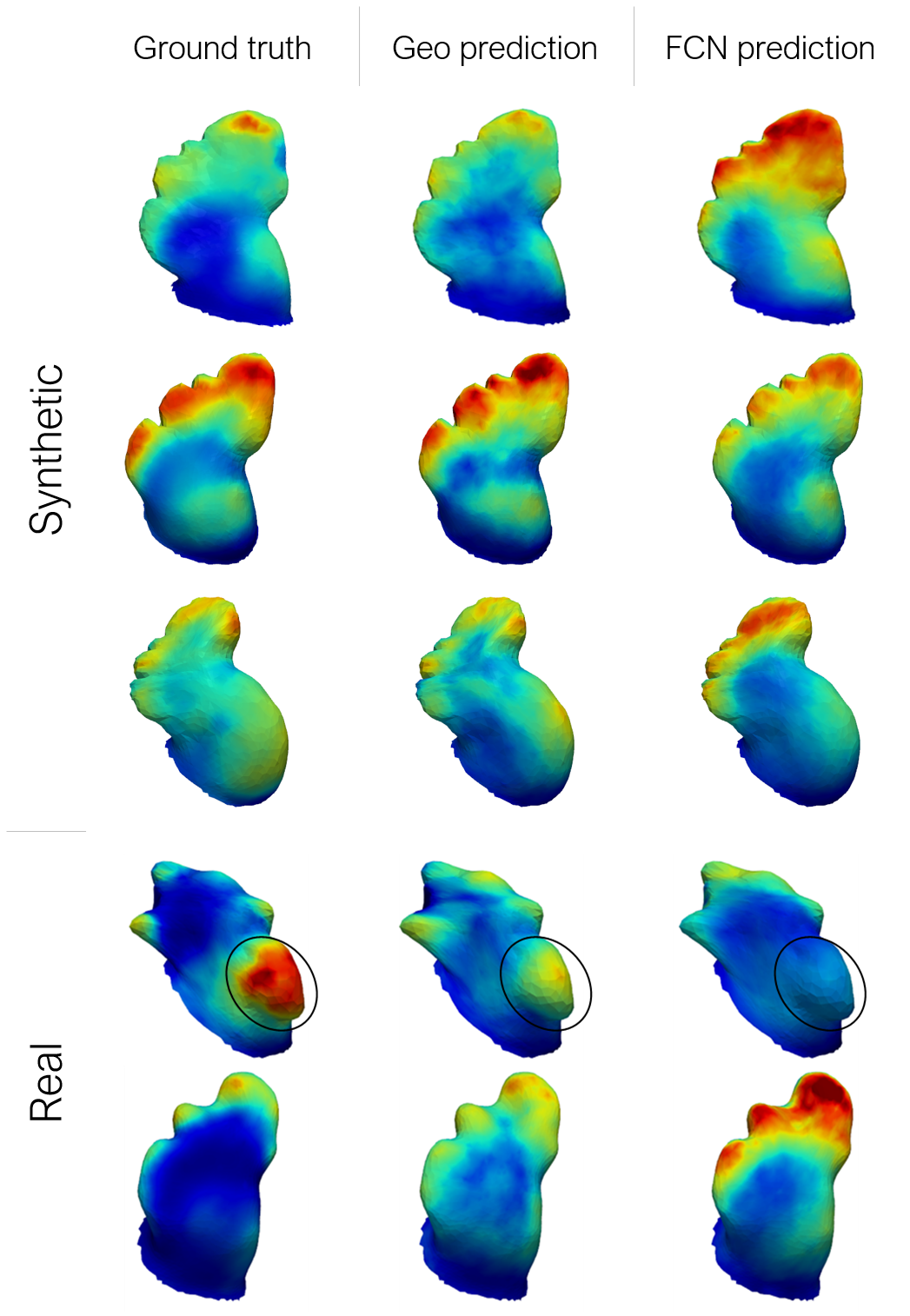}
\caption{From left to right: in-silico index (endothelial activation cell potential, ECAP) ground-truth from fluid simulations; prediction obtained with the geometrical deep learning model (Geo); and prediction from the fully connected network (FCN) \cite{Morales2020}. The ECAP values are colored from low values (0, blue) to higher than 6 (red areas), the latter indicating a higher risk of thrombus formation. The FCN model struggles to identify the highlighted lobe in the circle.}
\label{fig:Graphic Result}
\end{figure}

The accuracy results with the final model are given in Table \ref{table:Result Table} in terms of the mean absolute error (MAE), which indicates that the geometric DL network significantly outperforms the conventional fully connected network in both tasks. Furthermore, a small batch of 5 testing geometries from the first experiment is shown in Figure \ref{fig:Graphic Result}. Cases in row 1-3 are derived from the SSM model while the remaining two represent real patient cases. Additional test subjects are provided in Appendix \hyperref[appendix1]{A.1} and \hyperref[appendix2]{A.2}. As only the areas of high ECAP values are said to be related to increased risk of thrombosis, a binary classification was performed with a positive condition of ECAP $>$ 4, being the 90\textsuperscript{th} percentile of the distribution. Once again, the Geometric DL model outperformed its counterpart with a true positive rate of 73.1\% against 67.5\% in the FCN model.

\begin{table}
\includegraphics[width=1\textwidth]{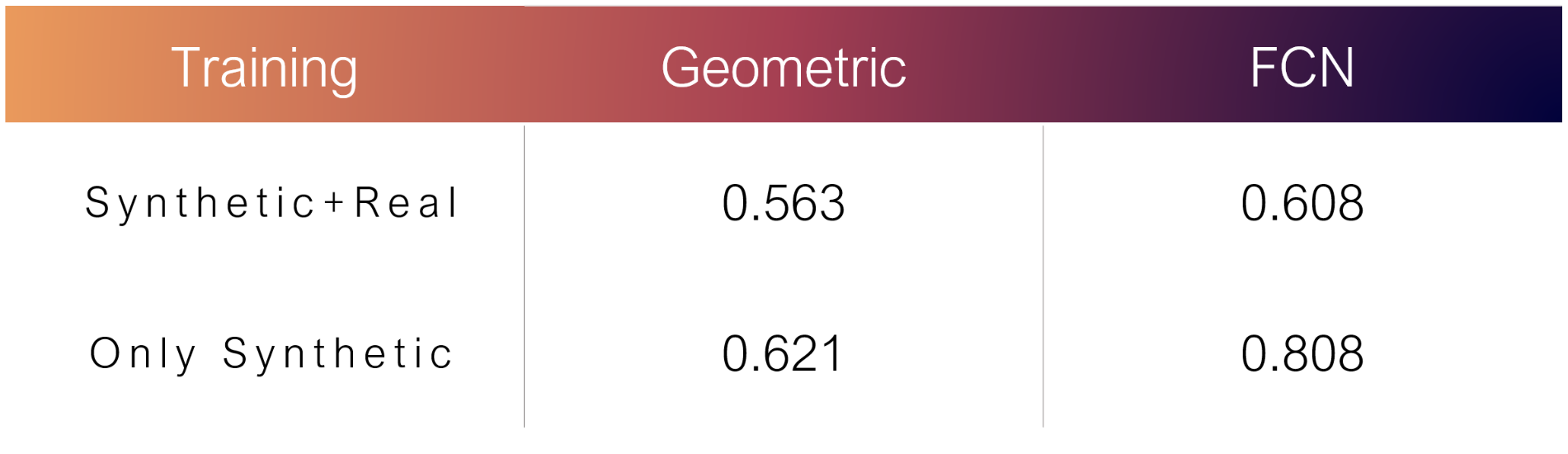}
\caption{Prediction accuracy results in terms of mean absolute error (MAE) for different training setups. FCN: Fully connected network. }
\label{table:Result Table}
\end{table}

\section{Discussion}

Careful inspection of the results in Table \ref{table:Result Table} indicate not only that the geometric DL model outperforms the conventional network but also that it has a higher generalisation potential. While the accuracy of the graph-based network decreases by just 9 \% when training solely in the synthetic data, the accuracy in the latter falls by almost 30\%. The drop in accuracy was to be expected as the real geometries present far higher heterogeneity than its synthetic counterparts. In this sense, the inclusion of a weight decay turned out be crucial in avoiding over-fitting the model to the synthetic cases.

Our hypothesis, although difficult to ascertain due to the "black box" nature of neural networks, is that the graph CNN model is probably better able to exploit the anatomical features in the vicinity of each node and, consequently, is capable of predicting higher ECAP values in areas with fluctuating curvature and normal vectors, which reflect the lobes and cavities of the LAA where blood tends to stagnate. Therefore, even though the network has only been provided with synthetic geometries during the training process, when tested on real cases it is able to recognise anatomical features such as bulges and gaps more proficiently, ultimately leading to improved accuracy. This is best exemplified in case 4 shown in Figure \ref{fig:Graphic Result}, as the FCN network completely fails to recognise the bulge (encircled in the figure), being in a region where the synthetic population rarely shows high ECAP values, while the graph-based network shows moderate success.

In spite of the results, this study has several limitations that must be addressed before it can be of any use in a clinical setting. First, the choice of ECAP as an index of thrombosis risk may be debatable, as its validity in the LAA has not yet been demonstrated in any clinical study. Nonetheless, although the ECAP index was originally developed in carotid and abdominal aorta fluid models \cite{DiAchille2014}, the underlying mechanisms of thrombus formation are analogous to those in the LAA, which typically involve some degree of blood stagnation or re-circulation at low velocities that the ECAP should be able to reflect. In fact, it has already seen some use in clinical studies exploring device-related thrombus formation in LAA occlusion surgeries \cite{Mill2020,Aguado2019}.

Secondly, the hemodynamic variability arising from the heterogeneous anatomy of the LA has been completely neglected for the sake of simplicity. Nonetheless, since the chosen deep learning framework does not involve mesh correspondence it should be fairly trivial to include the complete LA anatomy. Moreover, the network should be capable of learning the ECAP fluctuations caused by factors such as the interaction of pulmonary vein orientation \cite{Garcia-Isla2018}.

Lastly, at the moment, the model is completely agnostic to flow dynamics and boundary conditions that play a key role in the process of thrombogenesis. To address this challenge, we intend on capitalising on the rapid advances in the field of physics-informed neural networks, with examples such as the study by Pfaff et al. \cite{pfaff2021learning}, enabling the full exploration 4D flow MRI and CFD data that may pave the way towards the prediction of the velocity vector field in the LA.


\section{Conclusion}

In the present study we have successfully leveraged recent advances in graph neural networks to instantaneously predict the ECAP mapping in the LAA, solely from its anatomical mesh, effectively skipping the need to run CFD simulations. Furthermore, we have significantly improved the results from our previous model with a framework that no longer requires mesh correspondence. These results could lay the foundation for real-time monitoring of LAA thrombosis risk in the future and open exciting avenues for future research in cardiological mesh data. 

\subsubsection{Funding} This work was supported by the Agency for Management of University and Research Grants of the Generalitat de Catalunya under the the Grants for the Contracting of New Research Staff Programme - FI (2020 FI\_B 00608) and the Spanish Ministry of Economy and Competitiveness under the Programme for the Formation of Doctors (PRE2018-084062), the Maria de Maeztu Units of Excellence Programme (MDM-2015-0502) and the Retos Investigación project (RTI2018-101193-B-I00). Additionally, this work was supported by the H2020 EU SimCardioTest project (Digital transformation in Health and Care SC1-DTH-06-2020; grant agreement No. 101016496).

\subsubsection{Conflict of Interest Statement} The authors declare that the research was conducted in the absence of any commercial or financial relationships that could be construed as a potential conflict of interest.

\newpage


\section*{Appendix}
\subsection*{A.1 Real LAA results}\label{appendix1}
\begin{figure}[H]
\includegraphics[width=1\textwidth]{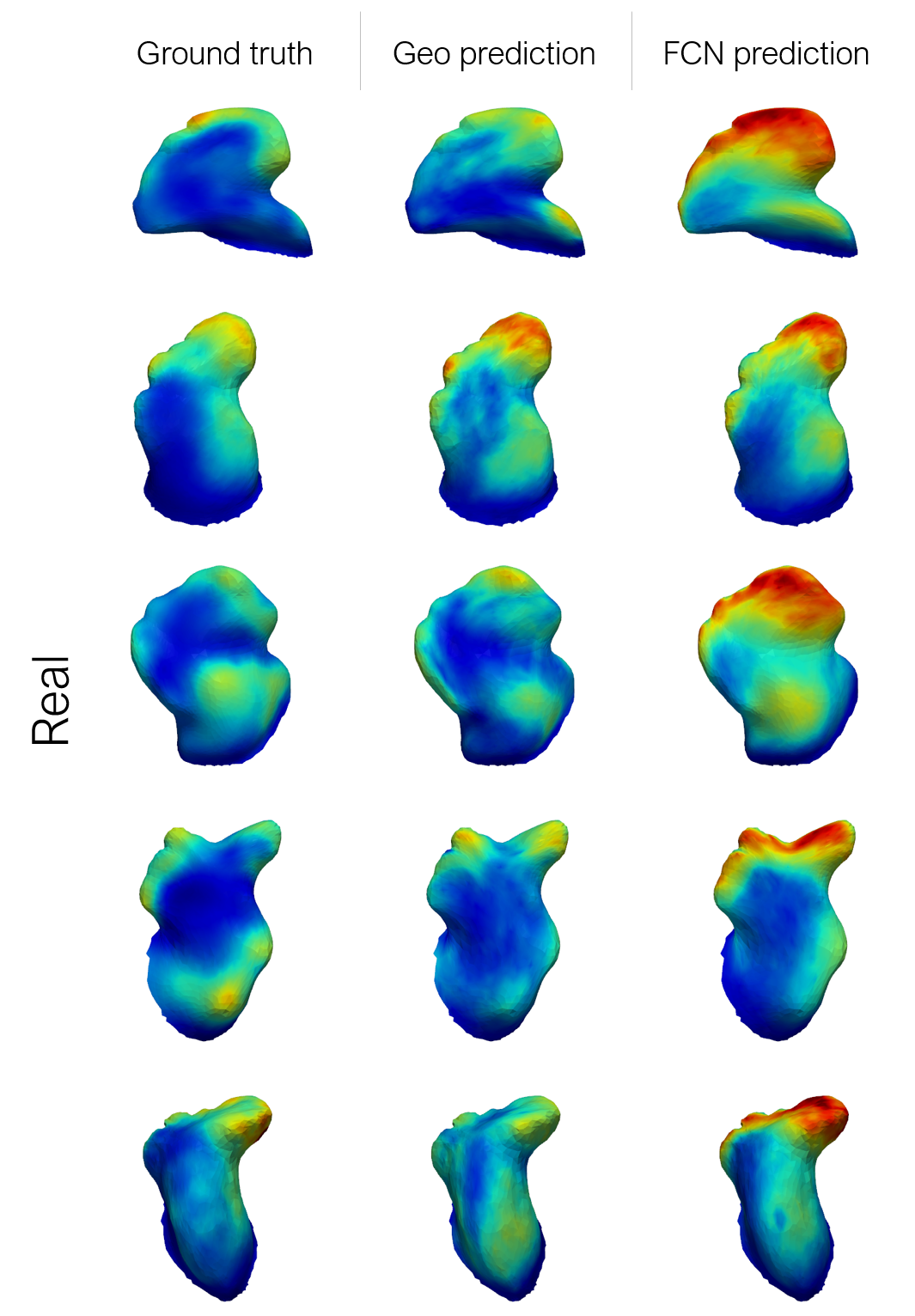}
\caption{Predictions over real LAA, belonging to the same cross-validation experiment as Figure \ref{fig:Graphic Result}. From left to right: in-silico index (endothelial activation cell potential, ECAP) ground-truth from fluid simulations; prediction obtained with the geometrical deep learning model (Geo); and prediction from the fully connected network (FCN) \cite{Morales2020}. The ECAP values are colored from low values (0, blue) to higher than 6 (red areas), the latter indicating a higher risk of thrombus formation.}
\label{fig:Real result}
\end{figure}

\subsection*{A.2 Synthetic LAA results}\label{appendix2}
\begin{figure}[H]
\includegraphics[width=1\textwidth]{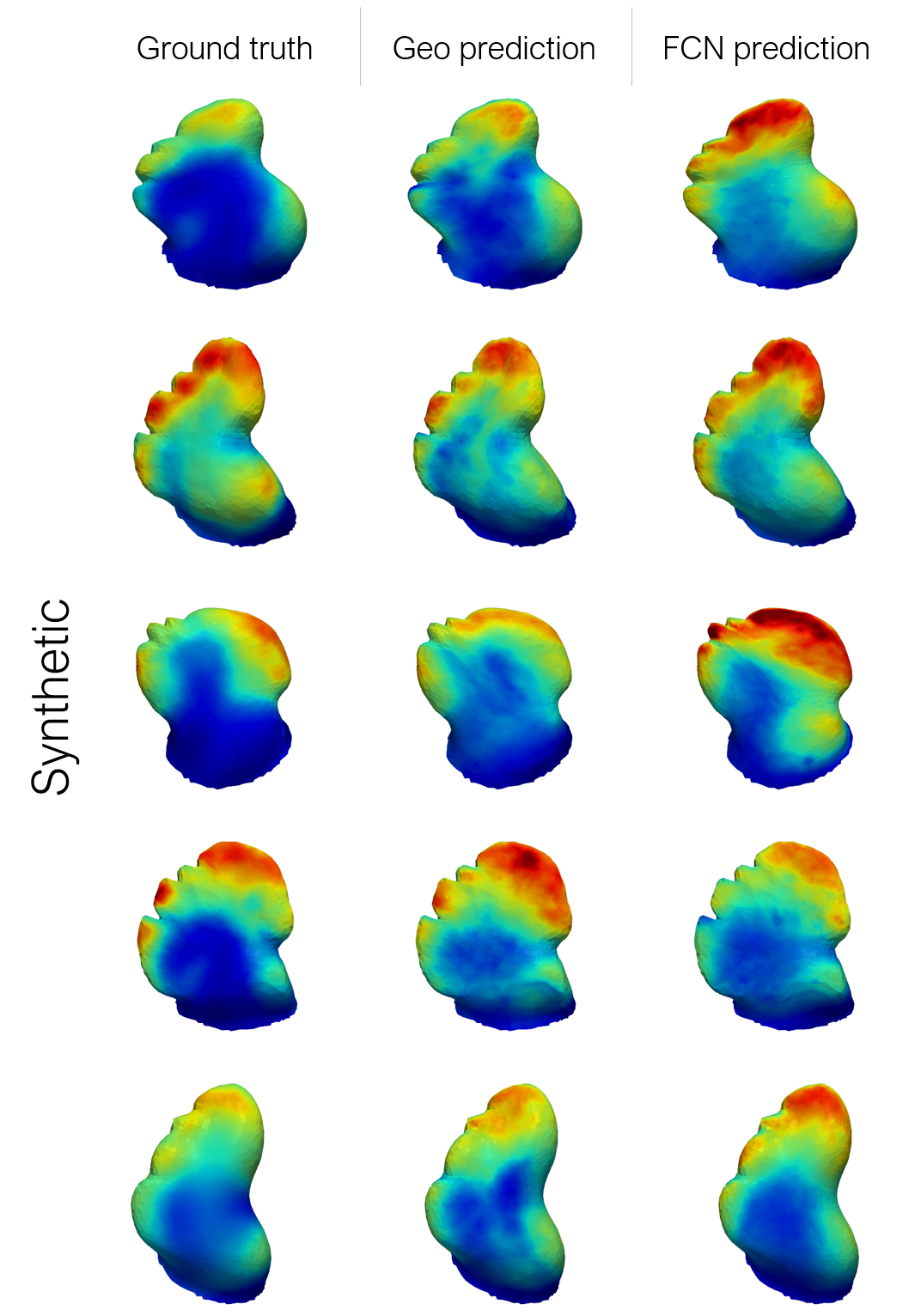}
\caption{Predictions over synthetic LAA, belonging to the same cross-validation experiment as Figure \ref{fig:Graphic Result}. From left to right: in-silico index (endothelial activation cell potential, ECAP) ground-truth from fluid simulations; prediction obtained with the geometrical deep learning model (Geo); and prediction from the fully connected network (FCN) \cite{Morales2020}. The ECAP values are colored from low values (0, blue) to higher than 6 (red areas), the latter indicating a higher risk of thrombus formation.}
\label{fig:Synthetic result}
\end{figure}


\bibliographystyle{splncs04}
\bibliography{bibliography}

\clearpage


\end{document}